\documentclass[fleqn,10pt,twocolumn]{ICCAS2020}

\usepackage{cite}
\usepackage{graphicx}
\usepackage{amsmath, amssymb}
\usepackage{booktabs}
\usepackage{multirow}
\usepackage{multicol}
\usepackage{hyperref}
\hypersetup{ colorlinks=true, linkcolor=red, filecolor=magenta, urlcolor=magenta}
\usepackage[center]{caption}

\DeclareMathOperator{\E}{\mathbb{E}}

\begin{document}

\title{Gaussian RAM: Lightweight Image Classification via
Stochastic\\ Retina-Inspired  Glimpse and Reinforcement Learning}

\author{Dongseok Shim${}^{1}$ and H. Jin Kim${}^{1*}$ }

\affils{ ${}^{1}$Department Mechanical \& Aerospace Engineering, Seoul National University, \\
Seoul, 08826, Korea (tlaehdtjr01@snu.ac.kr, hjinkim@snu.ac.kr) ${}^{*}$ Corresponding author}


\abstract{Previous studies on image classification have mainly focused on the performance of the networks, not on real-time operation or model compression. We propose a Gaussian Deep Recurrent visual Attention Model (GDRAM) - a reinforcement learning based lightweight deep neural network for large scale image classification that outperforms the conventional CNN (Convolutional Neural Network) which uses the entire image as input. Highly inspired by the biological visual recognition process, our model mimics the stochastic location of the retina with Gaussian distribution. We evaluate the model on Large cluttered MNIST, Large CIFAR-10 and Large CIFAR-100 datasets which are resized to 128 in both width and height. The implementation of Gaussian RAM in PyTorch and its pretrained model are available at :
 \texttt{\href{https://github.com/dsshim0125/gaussian-ram}{https://github.com/dsshim0125/gaussian-ram}}}

\keywords{
    Deep Learning, Computer Vision, Image Classification, Reinforcement Learning
}

\maketitle


\section{Introduction}
Computer vision is one of the most widely studied and strikingly developed fields due to the success of deep neural networks especially in image classification tasks\cite{Lecun98gradient-basedlearning, simonyan2014very,szegedy2015going, he2016deep, huang2017densely} from the early implementation of the machine learning in computer vision task using MNIST datasets\cite{lecun1998mnist} to large scale ImageNet visual recognition challenge (ILSVRC) \cite{ILSVRC15}.
 Although the winners of ILSVRC \cite{simonyan2014very, szegedy2015going, he2016deep} achieved state of the art performance, even outperforming humans, they have a myriad of parameters in  their networks (VGG19\cite{simonyan2014very}:145M, Inception-v3\cite{szegedy2015going}:27M, ResNet50\cite{he2016deep}:25M, DenseNet169\cite{huang2017densely}:15M). There exists a trade-off between model complexity and performance, so image classification networks have been developed to increase the model parameters or connection between layers for the performance of deep neural networks. However, this way of development discourages the implementation of high-quality image classification networks on local devices such as mobile robots or drones.
 
 In this paper, we propose a lightweight deep neural network (the number of the parameters of the network is about 1M) for large-scale image classification which outperforms conventional 2 layer CNN networks with the same number of parameters.
 
 We apply a glimpse network first suggested in RAM network\cite{mnih2014recurrent}. Humans recognize image not by using the entire image but rather leveraging only small spatial regions of the image. This small visual attention area for recognizing the entire image is called \textit{glimpse}. Based on the previous glimpse, humans decide where to look at next and the entire sequence of the glimpses is used classify the image. RAM and our method model this sequential policy with recurrent neural networks(RNN) and train them with reinforcement learning.
 
 Unlike the previous methods\cite{mnih2014recurrent, ba2014multiple}, we include the location of the glimpse to the state space and its movement to the action space which is more suitable for the dynamics model in the reinforcement learning and model the location of the glimpse with Gaussian distribution to estimate the uncertainty. The uncertainty works as a weight of each glimpse to give a different priority when classifying the image. This uncertainty-based weights give advantages in the perspective of the performance without increasing large numbers of parameters. We set a criterion to stop the glimpse sequence early so that the inference time can be shortened without degrading its performance much. In short, our contribution can be summarized as below.\\
 
 \begin{itemize}
     \item Redefine Markov model which is more suitable for  the dynamics model of the environment.
     \item Achieve higher accuracy by modeling the location of the glimpse with Gaussian distribution.
     \item Fast inference with uncertainty based early stopping.\\
 \end{itemize}
 
 We evaluate our Gaussian RAM on Large cluttered MNIST, Large CIFAR-10, and Large CIFAR-100. Our method outperforms the 2-layer conventional CNN, RAM \cite{mnih2014recurrent} and DRAM\cite{ba2014multiple} on top-1 and top-5 classification accuracy by a large margin.
 
\begin{figure*}[t]
\centering
\includegraphics[scale=0.45]{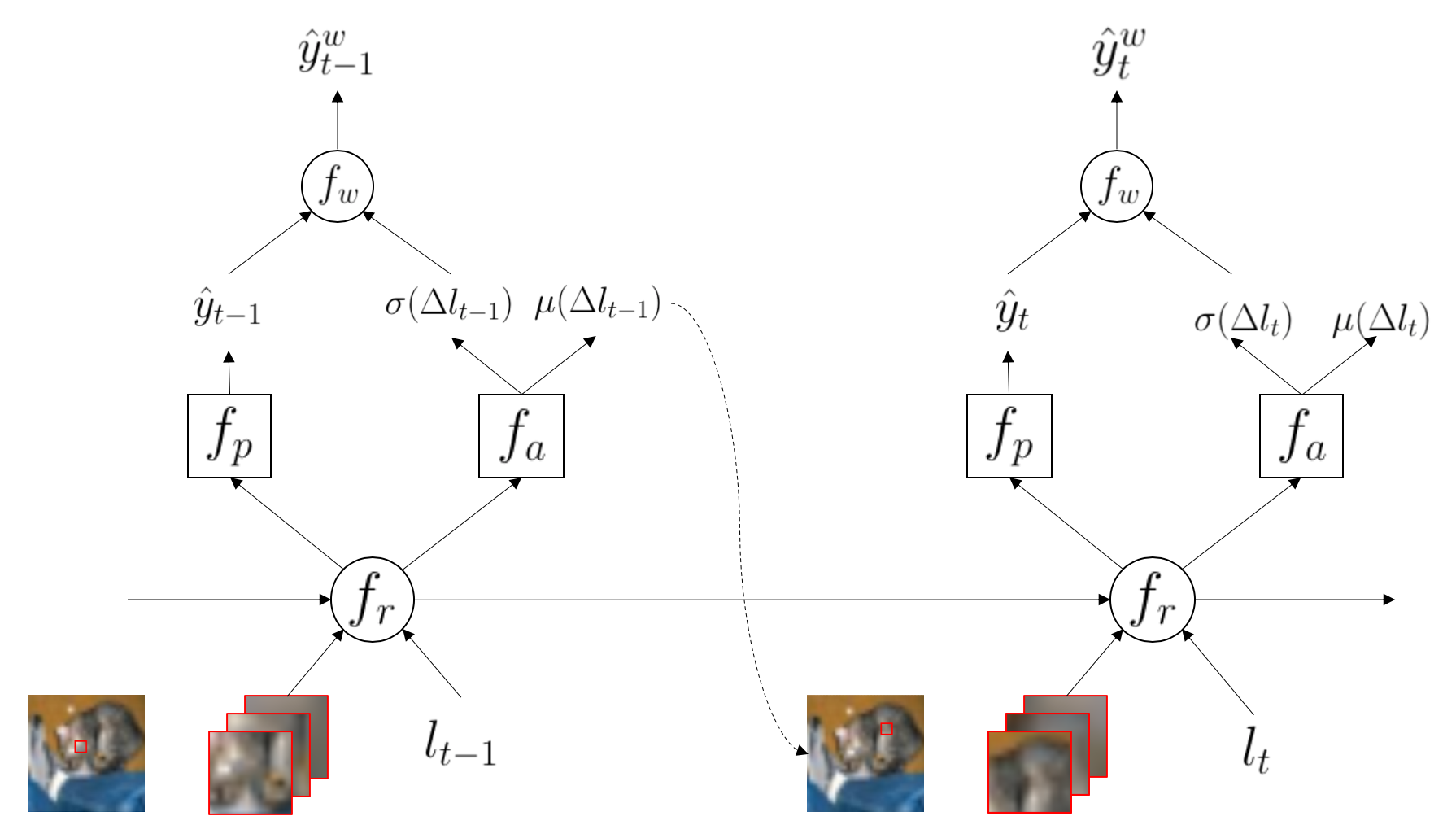}
\caption{Gaussian Recurrent Visual Attention Model}
\label{fig1}
\end{figure*}

\section{Related Work}
\subsection{High Resolution Image Classification}
There have been studies on large-scale (or high resolution) image classification \cite{liu2016learning, ifteneavery}. They either use the entire image without resizing \cite{ifteneavery} or resized image to multiple scales\cite{liu2016learning} as the input of the networks. However, if they use the entire image, the model parameters will increase exponentially and if they resize the input image, the resolution or the aspect ratio of the image will be distorted. Both cases might lead to low performance of the network so the invariance to input image size becomes the condition of a better model structure for image classification.
\subsection{Glimpse Network}
 A retina-inspired glimpse network has first been introduced in RAM \cite{mnih2014recurrent} which concatenates multiple resolution of glimpses at a single timestep. When humans look at a specific point, they can get information not only from the focused region but also from the point out of the attention area due to the biological structure of the retina. RAM doubles the glimpse size in multiple times and resize them back for the diversity of the resolution and semantic information which we call a \textit{retina-like} structure. It concatenates these glimpses with the same spatial size to use as the input of the networks.
 
 There were several follow-up studies of RAM including DRAM \cite{ba2014multiple} which aims to classify multi-labeled images with a glimpse network. RAM uses the same RNN hidden variables to predict image labels and next glimpse location coordinates. DRAM insists that this structure causes information loss when extracting hidden variables so it adds another RNN layer to separate the information flow that each RNN cell needs to capture. This structure leads to better performance compared to RAM in multiple object recognition but increases the model parameters (almost doubled). EDRAM\cite{ablavatski2017enriched} has a similar structure to DRAM but it applies spatial transformer\cite{jaderberg2015spatial} for homography transformation invariance.
 
\subsection{Gaussian Modeling}
In object detection tasks, YOLO-v3\cite{redmon2018yolov3} has widely been used for its high inference speed and adequate performance. The network predicts both the class score and the bounding box coordinates but because the bounding box is deterministic, the network does not know the uncertainty of the estimated bounding box. On the other hand, Gaussian YOLO-v3\cite{choi2019gaussian} models the bounding box with Gaussian distribution so that the standard deviation of the bounding box could indicate the uncertainty of the predictions. It proposes a new criterion in order to utilize the localization uncertainty during both training and inference step with the combination of object score, class score, and standard deviations.

\section{Method}
\subsection{Markov Model}
We redefine the Markov model compared to the previous variations of RAM \cite{mnih2014recurrent} so that it can be more well fitted to the dynamics of the environment for reinforcement learning.

We set the state of the environment with the original image $I$, multiple scales of the glimpses on the original image with the same spatial size $g_{t}$, and the location of the glimpses $l_{t}$ at the $t$ glimpse step.

Unlike the previous methods which use the location of the glimpses directly as the action of the policy, we include the location $l_{t}$ in the state of the environment and set the action as the movement of the eye glimpses or the difference of the location $\Delta l_{t}$ and the prediction of the classes $\hat{y}_{t}$. We design the reward $r_{t}$ as 1 if our model well classifies the object or otherwise 0.
\begin{align}
\qquad \qquad \qquad& s_{t} = \{I, \;g_{t},\; l_{t}\}\\
\qquad \qquad \qquad& a_{t} = \{\Delta l_{t},\; \hat{y}_{t}\}\\
\qquad \qquad \qquad& r_{t} =
    \begin{cases}
      1 & \text{if \; $\hat{y}_{t} = labels$}\\
      0 & \text{otherwise}
    \end{cases}
\end{align}

\subsection{Network Structure}
Our model consists of 4 neural networks as shown in Fig.\ref{fig1} which are responsible for different tasks to capture the both temporal and spatial information through the glimpse sequence.
\subsubsection{Glimpse network}
 The glimpse network is generating the latent variable $z_{t}$ from the glimpses of the image and its locations. It extracts the image features $z_{image , t}$ from the glimpses $g_{t}$ which have various resolutions and semantic information of the original images. Glimpses with high resolution have small semantic information and glimpses with low resolution have large semantic information.
 
 All the glimpses are concatenated channel-wise and used as the input of the network. The glimpse network consists of 2 layer CNN with 2D batch normalization\cite{ioffe2015batch} followed by a fully-connected layer.
 
 The location $l_{t}$ passes through two-layer MLP (Multi-Layer Perceptron) to extract the spatial information of the glimpses $z_{loc, t}$.
 
 The outputs of the CNN and MLP have the same dimension and the latent variable $z_{t}$ is generated by the multiplication of two feature extractions $z_{image, t}$, $z_{loc, t}$ so that it $z_{t}$ can represent both semantic and spatial information of the glimpses.
 \begin{equation}
 \centering
\qquad \qquad \qquad \quad z_{t} = z_{image, t} \times z_{loc, t}
 \end{equation}
 \subsubsection{Recurrent network}
Recurrent neural networks (RNN) are applied to capture the temporal information during the glimpse procedure. We fix the length of the glimpse $T = 8$ (or the length of the episode in RL style). Similar to \cite{ba2014multiple}, we implement two LSTM\cite{hochreiter1997long} layers $f_{r}^{1}, f_{r}^{2}$ to divide the information flow so that each RNN layer could capture the spatial information of the glimpse location and the semantic information of the image respectively for long-term range. Lower-layer RNN captures the semantic information and higher layer captures the spatial information. Hidden variables and cell states $h_{t}, c_{t}$ transfer the information from 
the previous glimpses to the next time step of each RNN cell.
\begin{gather}
\qquad \qquad \qquad  h_{t}^{1},\, c_{t}^{1} = f_{r}^{1}(z_{t}, h_{t-1}^{1}, c_{t-1}^{1})\\
\qquad \qquad \qquad  h_{t}^{2},\, c_{t}^{2} = f_{r}^{2}(h_{t}^{1}, h_{t-1}^{2}, c_{t-1}^{2})
\end{gather}

\subsubsection{Action network}
Action network $f_{a}$ leverages hidden variable $h_{t}^{2}$ of higher level RNN cells as the input. Networks consist of two layers of MLP and the output is the difference of the location of the glimpse $\Delta l_{t}$. However, the network does not na\"ively return the location difference but we model it as Gaussian distribution so the output would be stochastic. It leads the model to control the exploration-exploitation trade-off ratio adaptively by itself.
\begin{gather}
\qquad \qquad \quad     \mu (\Delta l_{t}) ,\; log(\sigma^{2} (\Delta l_{t})) = f_{a}(h_{t}^{2})\\
\qquad \qquad \qquad \Delta l_{t} \sim {\mathcal{N}} (\mu(\Delta l_{t}), \, \sigma^{2}(\Delta l_{t}))
\end{gather}

We limit the range of $\Delta l_{t}$ to smaller than the half of the maximum scale of the glimpse (width: $w_{g}$, height: $h_{g}$). It seems plausible that we should decide the next glimpse based on the information that we have seen especially from the previous glimpses. In this paper, we assume the shape of the glimpse as the square ($w_{g} = h_{g}$).
\begin{equation}
\qquad \;\;\;\;\; \Delta l_{t} =
\begin{cases}
-w_{g}/2 & \text{if} \; \Delta l_{t} < -w_{g}/2\\
w_{g}/2  & \text{if} \; \Delta l_{t} > w_{g}/2 \\
\Delta l_{t}  & \text{otherwise}
\end{cases}
\end{equation}

\subsubsection{Prediction network}
 The structure of the prediction network $f_{p}$ is quite similar to the action network in that it also consists of 2 layer MLP and uses the hidden variables $h_{t}^{1}$ of the lower level RNN as the input, but the shape of the output is same as the number of the class or object that the network needs to classify.
 \begin{equation}
 \qquad \qquad \qquad \qquad \quad \hat{y}_{t} = f_{p}(h_{t}^{1})
 \end{equation}
\subsection{Uncertainty Weighted Prediction}
The benefits from modeling the glimpse location with Gaussian distribution are first, the model can adaptively behave exploration and exploitation for policy robustness to noise. Second, we can estimate the uncertainty of the current glimpse location via its standard deviation. In the sequence of a single classification procedure, multiple glimpses are given and each glimpse generates the prediction of the class with the deep neural networks. Previous methods consider all the glimpses equally so they calculate the final result of the classification as the average of every glimpse. We introduce the weight of the glimpse to apply the uncertainty to the classification result. The outputs of the action network which are the mean and log variance of the location difference $\Delta l$ are the result of the hyperbolic-tangent(tanh) operation. So, the in range would be from -1 to +1 and we can calculate the minimum and maximum of the standard deviation.
\begin{gather}
\qquad \qquad \qquad -1 \leq log(\sigma ^{2}(\Delta l_{t})) \leq 1\\
\qquad \qquad \qquad \; \; \; e^{-\frac{1}{2}} \leq \sigma(\Delta l_{t}) \leq e^{\frac{1}{2}}
\end{gather}

 We normalize the standard deviation of the location difference from 0 to 1 and this indicates the uncertainty of the location of the glimpse. Weight for each glimpse is defined as Eq. (14) and the final classification output is calculated as the weighted average of each glimpse.
\begin{gather}
\qquad \qquad \qquad \sigma_{norm} = \frac{\sigma (\Delta l_{t}) - e^{-\frac{1}{2}}}{e^{\frac{1}{2}} - e^{-\frac{1}{2}}}\\
\qquad \qquad \qquad\quad w_{t} = 1 - \sigma_{norm}\\
\qquad \qquad \qquad \quad \hat{y}_{t}^{w} = w_{t} \times \hat{y}_{t}\\
\qquad \qquad \qquad \quad \hat{y}^{w} = \frac{\Sigma_{t=0}^{T}\hat{y}_{t}^{w}}{\Sigma_{t=0}^{T}w_{t}}
\end{gather}

\begin{table*}[t]
\centering
\begin{tabular}{c|c|c|c|c|c|c|c}
&Model  & 2-layer CNN &RAM & RAM w/ $\Delta l$ & GRAM &DRAM&Ours\,(GDRAM)\\

\midrule[1.3pt]

Large CIFAR-10& top-1&  48.55 & 55.24 & 59.94&63.27&62.29&\textbf{66.44}\\

\hline
Large Clutterd MNIST& top-1&  65.20&76.20 & 76.40 & 85.20&71.20&\textbf{95.20}\\

\hline
\multirow{2}{*}{Large CIFAR-100} &top-1&18.68&23.80&26.13&28.94&21.08&\textbf{33.90}\\
                              &top-5&41.32&51.28&54.36&58.15&46.77&\textbf{63.15}\\
\midrule[0.7pt]
                      &params&  1.57M & 1.30M & 1.30M&1.30M&1.13M&1.13M
\end{tabular}
\captionsetup{justification=raggedright,singlelinecheck=false}
    \caption{Classification accuracy (\%) on Large Cluttered MNIST, Large CIFAR-10 and Large CIFAR-100. All the RAM\cite{mnih2014recurrent} variation networks use 4 glimpses with 12$\times$12 size on a single timestep. RAM w/ $\Delta l$ indicates the application of the newly defined Markov model on RAM. Prefix G shows that the network models the location of the glimpse as Gaussian distribution. The parameters of the model differ according to the dataset due to the number of the object classes and in this table, we calculate the parameters with 10 classes.}
    \label{tab:1}
\end{table*}

\begin{figure*}[t]
\centering
\includegraphics[scale=0.60]{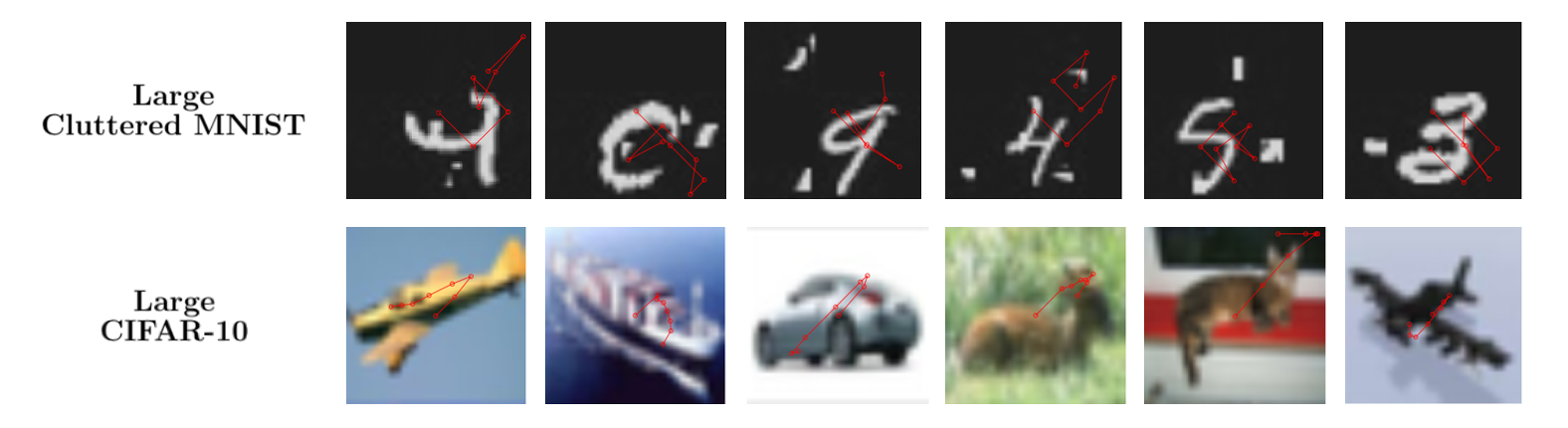}
\captionsetup{justification=raggedright,singlelinecheck=false}
\caption{Visualization of the glimpse sequence for image classification. Each vertex and edge indicates the attention point of the glimpse and its trajectory respectively.}
\end{figure*}

\subsection{Training}
 The policy of our RL setting $\pi _{\theta}$ consists of two different neural networks $f_{p}$, $f_{a}$ that generate different outputs and different loss functions.
 
 We train action network $f_{a}$ to maximize the expected return $J_{\theta}$ from each glimpse trajectory $\tau$ which is the objective of the trivial reinforcement learning.
\begin{gather}
    \qquad \qquad \qquad \theta^{*} = \mathop{argmax}_{\theta} J(\theta) \\
    \qquad \qquad \quad J(\theta) = \E _{\tau \sim \pi_{\theta}(\tau)}[r(\tau)]
\end{gather}
 
The gradient of the objective function $J(\theta)$ can be simplified by REINFORCE algorithm \cite{sutton2000policy} and for the unbiased expectation, we additionally train baseline networks $f_{b}$ with two-layer MLP to follow the reward $r_{t}$. The input of the baseline networks is given as the concatenation of the two hidden variables from the higher and lower level of recurrent networks.
\begin{gather}
\; \qquad \qquad \qquad \quad h_{t} = cat(h_{t}^{1}, h_{t}^{2})\\
    \qquad \qquad \qquad \qquad b_{t} = f_{b}(h_{t})\\
    \;\;\qquad \qquad \qquad \; L_{b} = \Sigma_{t}||b_{t} - r_{t}||^{2}\\
    \;\; \quad \nabla J(\theta) = \E_{\tau \sim \pi_{\theta}(\tau)}[\nabla_{\theta}log\pi_{\theta}(\tau)(r(\tau) - b)]\\
    \;\; \qquad \qquad \approx \Sigma_{i}(\Sigma_{t}\nabla_{\theta}log\pi_{\theta}(a_{t}^{i}|s_{t}^{i}))(\Sigma_{t}(r_{t}^{i} - b_{t}^{i}))
\end{gather}

 The final class prediction of our model $\hat{y}^{w}$can be formulated as the linear combination of the output of prediction network $f_{p}$ and action network $f_{a}$. We train both networks as supervised learning to reduce the cross-entropy between final prediction and class label.
 \begin{equation}
     \qquad \qquad \qquad \qquad L_{p} = D_{KL}(\hat{y}^{w} || y)
 \end{equation}
 
 So, the network is trained with three losses $L_{p}$, $L_{RL}$, $L_{b}$ for predicting both the location of the glimpses and the class of the image.

\section{Experimental Results}
\subsection{Dataset}
We evaluate our model with Large Cluttered MNIST (LCM), Large CIFAR-10 (LC10), and Large CIFAR-100 (LC100) datasets which we intentionally upsample the size of the images to 128 in both width and height to check the model performance on large-scale image.
\subsection{Ablation Studies}
We compare conventional 2-layer CNN, RAM\cite{mnih2014recurrent}, DRAM\cite{ba2014multiple}, and the applications of our proposed methods with a similar number of model parameters for fair evaluation. We measured both top-1 and top-5 accuracy on Large CIFAR-100 and top-1 accuracy for Large CIFAR-10, Large Cluttered MNIST.

\noindent\textbf{Markov model} To evaluate the difference in model performance by re-designing the Markov model to include the location of the glimpse to the environment state and location movements to action space, we train two RAMs\cite{mnih2014recurrent} with the same model structure and different Markov model. Column 4, 5 of Table \ref{tab:1} show that there has been $2\%\sim3\%$ increase in the accuracy, especially for Large CIFAR-10 dataset with 4.7\% developments.

\noindent\textbf{RAM vs DRAM} DRAM has a similar network structure to RAM but splits the information flow by adding another RNN layer. In the original model DRAM has almost twice the number of parameters compared to RAM but in this paper, we reduce all the number of hidden parameters of DRAM by half to make two networks have a similar model complexity. Columns 5 and 7 of Table \ref{tab:1} show that RAM outperforms the compressed DRAM in Large CIFAR-100 and Large Cluttered MNIST but compressed DRAM performs better in Large CIFAR-10.

\noindent\textbf{Gaussian modeling} Columns 6 and 8 of Table \ref{tab:1} indicate the networks with Gaussian modeling of glimpse locations. The difference in model structure is only at the last layer of the action network due to the output shape, so there is little change in model parameters. Gaussian-RAM(GRAM) shows a performance increase 4.68\% and Gaussian-DRAM(GDRAM) 14.33\% in average of top-1 and top-5 classification accuracy 
compared to the deterministic models (RAM, DRAM).

\subsection{Uncertainty based Inference}
By modeling the location of the glimpse with Gaussian distribution, we could estimate the uncertainty of each glimpse with its standard deviation $\sigma (\Delta l)$. As the final class prediction is calculated as the weighted average of the outputs of each glimpse, uncertainty based weight indicates the importance of the glimpse. We assume that the classification is done if the weights of the glimpse are below 0.5 twice in a row, and prematurely terminate the glimpse sequence for the inference time efficiency. We compare this early stopping model Fast-Gaussian DRAM (F-GDRAM) to full glimpse step ($T$=$8$) Gaussian DRAM (GDRAM). The inference time is tested on a single GPU, GEFORCE RTX 2080 Ti. In Table \ref{tab:2}, there exists a trade-off between classification accuracy and inference time per image. The accuracy is decreased about $3\%\sim5\%$ but the inference time is almost half and the accuracy of F-GDRAM still outperforms all the other RAM variations except GDRAM.
\begin{table}[t]
    \centering
    \begin{tabular}{c|c|c|c|c}
         &Dataset&\multicolumn{2}{c}{Accuracy}& Time\\
         \midrule[1.3pt]
         \multirow{4}{*}{GDRAM}&LCM& top-1& 95.20&33ms \\
         \cline{2-5}
         &LC10& top-1&66.44&29ms \\
         \cline{2-5}
         &\multirow{2}{*}{LC100}&top-1&33.90&\multirow{2}{*}{31ms}\\
         \cline{3-4}
         &                      &top-5&63.15&\\
         \hline
         \multirow{4}{*}{F-GDRAM}&LCM& top-1& 92.90&19ms \\
         \cline{2-5}
         &LC10& top-1&62.15&16ms \\
         \cline{2-5}
         &\multirow{2}{*}{LC100}&top-1&30.70&\multirow{2}{*}{15ms}\\
         \cline{3-4}
         &                      &top-5&58.67&
                               
    \end{tabular}
    \captionsetup{justification=raggedright,singlelinecheck=false}
    \caption{Accuracy (\%) and the inference time per image (ms) of GDRAM and uncertainty based early stopping method F-GDRAM.}
    \label{tab:2}
\end{table}


\section{Conclusion}
 In this paper, we propose a lightweight network for large-scale image classification with visual attention and Gaussian modeling. We redefine the Markov process for reinforcement learning and model the location of the glimpse with Gaussian distribution. With ablation studies with several datasets, we showed that our methods contribute to increasing the model performance and the uncertainty estimation with standard deviation works not only for increasing the classification accuracy by applying weighted average but also for decreasing the inference time.
\section{Acknowledgement}
This work was supported by Institute of Information \& Communications Technology Planning \& Evaluation(IITP) grant funded by the Korea government (MSIT) (No.2019-0-01367, Infant-Mimic Neurocognitive Developmental Machine Learning from Interaction Experience with Real World (BabyMind))

%
\bibliographystyle{IEEEtran}
\bibliography{main}

\begin{thebibliography}{10}
\providecommand{\url}[1]{#1}
\csname url@samestyle\endcsname
\providecommand{\newblock}{\relax}
\providecommand{\bibinfo}[2]{#2}
\providecommand{\BIBentrySTDinterwordspacing}{\spaceskip=0pt\relax}
\providecommand{\BIBentryALTinterwordstretchfactor}{4}
\providecommand{\BIBentryALTinterwordspacing}{\spaceskip=\fontdimen2\font plus
\BIBentryALTinterwordstretchfactor\fontdimen3\font minus
  \fontdimen4\font\relax}
\providecommand{\BIBforeignlanguage}[2]{{%
\expandafter\ifx\csname l@#1\endcsname\relax
\typeout{** WARNING: IEEEtran.bst: No hyphenation pattern has been}%
\typeout{** loaded for the language `#1'. Using the pattern for}%
\typeout{** the default language instead.}%
\else
\language=\csname l@#1\endcsname
\fi
#2}}
\providecommand{\BIBdecl}{\relax}
\BIBdecl

\bibitem{Lecun98gradient-basedlearning}
Y.~Lecun, L.~Bottou, Y.~Bengio, and P.~Haffner, ``Gradient-based learning
  applied to document recognition,'' in \emph{Proceedings of the IEEE}, 1998,
  pp. 2278--2324.

\bibitem{simonyan2014very}
K.~Simonyan and A.~Zisserman, ``Very deep convolutional networks for
  large-scale image recognition,'' \emph{arXiv preprint arXiv:1409.1556}, 2014.

\bibitem{szegedy2015going}
C.~Szegedy, W.~Liu, Y.~Jia, P.~Sermanet, S.~Reed, D.~Anguelov, D.~Erhan,
  V.~Vanhoucke, and A.~Rabinovich, ``Going deeper with convolutions,'' in
  \emph{Proceedings of the IEEE conference on computer vision and pattern
  recognition}, 2015, pp. 1--9.

\bibitem{he2016deep}
K.~He, X.~Zhang, S.~Ren, and J.~Sun, ``Deep residual learning for image
  recognition,'' in \emph{Proceedings of the IEEE conference on computer vision
  and pattern recognition}, 2016, pp. 770--778.

\bibitem{huang2017densely}
G.~Huang, Z.~Liu, L.~Van Der~Maaten, and K.~Q. Weinberger, ``Densely connected
  convolutional networks,'' in \emph{Proceedings of the IEEE conference on
  computer vision and pattern recognition}, 2017, pp. 4700--4708.

\bibitem{lecun1998mnist}
Y.~LeCun, ``The mnist database of handwritten digits,'' \emph{http://yann.
  lecun. com/exdb/mnist/}.

\bibitem{ILSVRC15}
O.~Russakovsky, J.~Deng, H.~Su, J.~Krause, S.~Satheesh, S.~Ma, Z.~Huang,
  A.~Karpathy, A.~Khosla, M.~Bernstein, A.~C. Berg, and L.~Fei-Fei, ``{ImageNet
  Large Scale Visual Recognition Challenge},'' \emph{International Journal of
  Computer Vision (IJCV)}, vol. 115, no.~3, pp. 211--252, 2015.

\bibitem{mnih2014recurrent}
V.~Mnih, N.~Heess, A.~Graves \emph{et~al.}, ``Recurrent models of visual
  attention,'' in \emph{Advances in neural information processing systems},
  2014, pp. 2204--2212.

\bibitem{ba2014multiple}
J.~Ba, V.~Mnih, and K.~Kavukcuoglu, ``Multiple object recognition with visual
  attention,'' \emph{arXiv preprint arXiv:1412.7755}, 2014.

\bibitem{liu2016learning}
Q.~Liu, R.~Hang, H.~Song, and Z.~Li, ``Learning multi-scale deep features for
  high-resolution satellite image classification,'' \emph{arXiv preprint
  arXiv:1611.03591}, 2016.

\bibitem{ifteneavery}
M.~Iftenea, Q.~Liub, and Y.~Wangc, ``Very high resolution images classification
  by fusing deep convolutional neural networks.''

\bibitem{ablavatski2017enriched}
A.~Ablavatski, S.~Lu, and J.~Cai, ``Enriched deep recurrent visual attention
  model for multiple object recognition,'' in \emph{2017 IEEE Winter Conference
  on Applications of Computer Vision (WACV)}.\hskip 1em plus 0.5em minus
  0.4em\relax IEEE, 2017, pp. 971--978.

\bibitem{jaderberg2015spatial}
M.~Jaderberg, K.~Simonyan, A.~Zisserman \emph{et~al.}, ``Spatial transformer
  networks,'' in \emph{Advances in neural information processing systems},
  2015, pp. 2017--2025.

\bibitem{redmon2018yolov3}
J.~Redmon and A.~Farhadi, ``Yolov3: An incremental improvement,'' \emph{arXiv
  preprint arXiv:1804.02767}, 2018.

\bibitem{choi2019gaussian}
J.~Choi, D.~Chun, H.~Kim, and H.-J. Lee, ``Gaussian yolov3: An accurate and
  fast object detector using localization uncertainty for autonomous driving,''
  in \emph{Proceedings of the IEEE International Conference on Computer
  Vision}, 2019, pp. 502--511.

\bibitem{ioffe2015batch}
S.~Ioffe and C.~Szegedy, ``Batch normalization: Accelerating deep network
  training by reducing internal covariate shift,'' \emph{arXiv preprint
  arXiv:1502.03167}, 2015.

\bibitem{hochreiter1997long}
S.~Hochreiter and J.~Schmidhuber, ``Long short-term memory,'' \emph{Neural
  computation}, vol.~9, no.~8, pp. 1735--1780, 1997.

\bibitem{sutton2000policy}
R.~S. Sutton, D.~A. McAllester, S.~P. Singh, and Y.~Mansour, ``Policy gradient
  methods for reinforcement learning with function approximation,'' in
  \emph{Advances in neural information processing systems}, 2000, pp.
  1057--1063.

\end{thebibliography}

%

\end{document}